\documentclass[10pt,twocolumn,letterpaper]{article}

\usepackage{cvpr}              

\usepackage[dvipsnames,table]{xcolor}
\usepackage{multirow}

\usepackage{booktabs}
\usepackage{amsmath}
\usepackage{tikz}
\usepackage{empheq}
\usepackage{graphicx}
\usepackage{float}

\definecolor{lightred}{rgb}{1,0.8,0.8}
\definecolor{lightyellow}{rgb}{1,1,0.9}
\definecolor{lightgreen}{rgb}{0.9,1,0.9}

\newcommand{\MODEL}{Neural Plenoptic Function}
\newcommand{\model}{neural plenoptic function}
\newcommand{\mm}{NeP}

\newcommand{\fields}{neural radiance fields}

\newcommand{\gcell}[1]{{\bf #1}}
\newcommand{\wcell}[1]{#1}
\newcommand{\xx}{\mathbf{x}}
\newcommand{\dd}{\mathbf{d}}

\newcommand{\nn}{\mathbf{n}}

\newcommand{\cc}{\mathbf{c}}

\newcommand{\sdf}{\text{SDF}}

\newcommand{\reL}{L}

\newlength{\imw}

\definecolor{cvprblue}{rgb}{0.21,0.49,0.74}
\definecolor{mylinkcolor}{HTML}{7D7FC5}

\usepackage{graphicx}
\usepackage[pagebackref,breaklinks,colorlinks,citecolor=cvprblue,linkcolor=cvprblue]{hyperref}

\def\paperID{2147} 
\def\confName{CVPR}
\def\confYear{2024}

\newcommand{\ryn}[1]{{\color[rgb]{0,0,0}{#1}}}

\title{
Inverse Rendering of Glossy Objects via \\ the Neural Plenoptic Function and Radiance Fields
}

\author{
Haoyuan Wang\textsuperscript{1}, \ \ \ 
Wenbo Hu\textsuperscript{2†}, \ \ \ 
Lei Zhu\textsuperscript{1}, \ \ \ 
Rynson W.H. Lau\textsuperscript{1†} \\
\textsuperscript{1}City University of Hong Kong \ \ \ \  
\textsuperscript{2}Tencent AI Lab \\
\footnotesize{† Joint corresponding authors}
}

\begin{document}


\twocolumn[{%
\renewcommand\twocolumn[1][]{#1}%
\maketitle

\setlength{\imw}{0.163\textwidth}
\vspace{-5mm}
\renewcommand{\tabcolsep}{2pt}
\centering
\begin{tabular}{cccccc}

\includegraphics[width=1.0\linewidth]{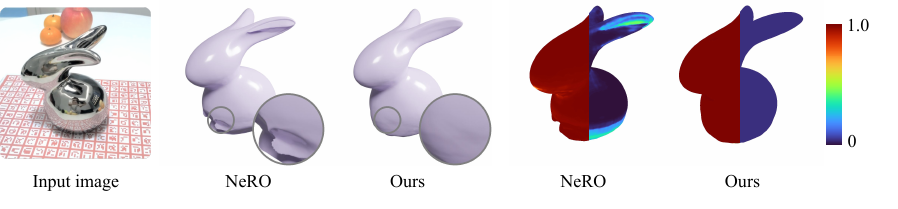}


\end{tabular}
\vspace{-2em}
\captionof{figure}{
Inverse rendering results of the cutting-edge method, NeRO~\cite{nero}, and ours from calibrated multi-view images of a glossy object.
Geometries are shown as a rendered mesh in the second and third images, and materials (metalness \& roughness) are shown as a color map in the fourth and fifth images.
We can see that our results not only have smoother and more accurate geometry but also 
present
a more reasonable material (since the material of this object should be uniform).   
}
\label{fig:teaser}
\vspace{5mm}
}]


\begin{abstract}
    \vspace{-3.5mm}
Inverse rendering aims at recovering both geometry and materials of objects. It provides a more compatible reconstruction for conventional rendering engines, compared with the neural radiance fields (NeRFs). On the other hand, existing NeRF-based inverse rendering methods cannot handle glossy objects with local light interactions well, as they typically oversimplify the illumination as a 2D environmental map, which assumes infinite lights only. Observing the superiority of NeRFs in recovering radiance fields, we propose a novel 5D Neural Plenoptic Function (NeP) based on NeRFs and ray tracing, such that more accurate lighting-object interactions can be formulated via the rendering equation. We also design a material-aware cone sampling strategy to efficiently integrate lights inside the BRDF lobes with the help of pre-filtered radiance fields. Our method has two stages: the geometry of the target object and the pre-filtered environmental radiance fields are reconstructed in the first stage, and materials of the target object are estimated in the second stage with the proposed NeP and material-aware cone sampling strategy. Extensive experiments on the proposed real-world and synthetic datasets demonstrate that our method can reconstruct high-fidelity geometry/materials of challenging glossy objects with complex lighting interactions from nearby objects. Project webpage: {\small \url{https://whyy.site/paper/nep}{}}


\end{abstract}    
\section{Introduction}
\label{sec:intro}

Although Neural Radiance Fields (NeRFs)~\cite{nerf,nerfpp,mipnerf,mipnerf360,ngp,trimipnerf,tensorf,zipnerf,plenoxel,rawnerf,merf,refnerf,f2nerf} have achieved remarkable progress in photo-realistic reconstruction, it is still a challenge to integrate NeRFs into conventional rendering engines since NeRFs represent the object and illumination in an entangled manner.
Disentangling the representation into geometry, materials, and environmental lighting, \textit{i.e.} inverse rendering, is crucial for the applicability in game production and extended reality.



%
%
%

%
Recent works have explored geometry reconstruction~\cite{neus,volsdf,neus2,unisurf,MonoSDF,neuralangelo,neuraludf} and further extended to the materials estimation~\cite{nvdiffrec, nvdiffrecmc, refnerf,nero,Zhang2021NeRFactor}, \eg, albedo, roughness, and metalness.
%
However, they typically represent the illumination as 2D environmental maps~\cite{nvdiffrec, nvdiffrecmc,nero}, which oversimplifies the complicated real-world lighting distribution to \emph{infinite lights} only. 
%
%
%
%
In many practical scenarios where the target object is surrounded by other objects, a considerable amount of light actually comes from the radiance of those nearby objects.
Neglecting these common scenarios results in inferior reconstruction of both geometry and materials, especially for glossy objects, such as the improper results of NeRO~\cite{nero} in Fig.~\ref{fig:teaser}.
%
%

In this paper, we propose a \textit{\MODEL~(\mm)} to represent the global illumination as a 5D function, $f_p(\xx, \dd)$, which describes the color of each light observed at position $\xx$ with direction $\dd$, in line with the definition of traditional plenoptic function~\cite{plenoptic}.
Observing the superiority of NeRFs in recovering radiance fields from multi-view images, we construct the \mm~from neural radiance fields based on a ray tracing procedure.
However, directly doing so is computationally intensive, because rendering a ray's color from NeRF via ray marching is expensive, and ray tracing requires sampling a large number of rays inside the BRDF lobe to approximate the integration in the rendering equation.
Thus, instead of sampling the lobe with rays, we propose an efficient material-aware cone-sampling strategy, where the cone's angle is derived from the predicted roughness.
The color of lights inside a cone can be directly rendered from pre-filtered radiance fields, thanks to the anti-aliasing techniques in Mip-NeRF~\cite{mipnerf}.

Overall, our method is divided into two stages: geometry reconstruction and material estimation of the target object.
In the first stage, our model consists of an {object field} with a decoupled color representation, \ie albedo and the color modulated by the lighting, and a pre-filtered environmental field for capturing scene radiance.
To promote high-quality geometry reconstruction for glossy objects, we design a dynamic weighting loss mechanism from the decoupled colors to reduce the impact of highly uncertain reflective regions while amplifying the significance of diffuse areas.
In the second stage, we adopt the Physically-Based Rendering (PBR) to estimate high-fidelity materials with our proposed \mm{} and material-aware cone-sampling strategy, based on the extracted triangular mesh of the target object and pre-trained environmental fields in the previous stage.
Our two-stage method can faithfully reconstruct both the geometry and material properties of the target object, as shown in Fig.~\ref{fig:teaser}, solely from calibrated multi-view images.
And importantly, the results can be seamlessly integrated into conventional rendering engines for 
relighting,
as our method can produce compatible triangle meshes with physically-based materials.

\begin{figure}
    \centering
    \includegraphics[width=1.05\linewidth]{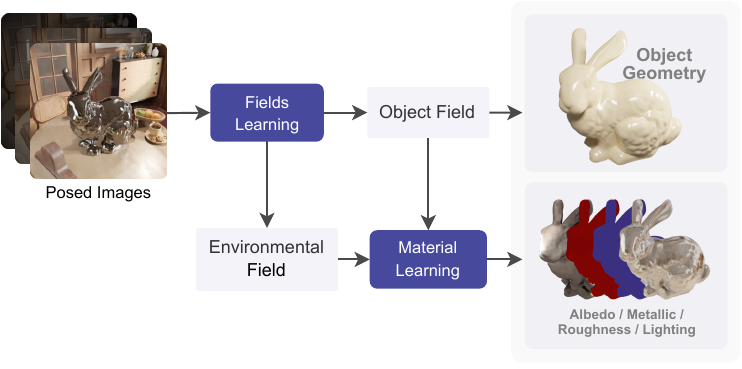}%
    \vspace{-3mm}
    \caption{
    The pipeline of the proposed method. Our method has two stages: the \textit{fields learning stage} for object geometry reconstruction and the {\fields} optimization, and the \textit{material learning stage} using ray tracing.
    }
    \vspace{-1.5em}
    \label{fig:pipeline}
\end{figure}

To evaluate our method, we compiled two challenging glossy object datasets, from the rendering engine and real-world captures, respectively.
%
Extensive experiments both quantitatively and qualitatively demonstrate that our method is robust, adaptable, and capable of handling diverse challenging illuminations.
%
%
We also demonstrate the possible applications of our reconstructions, \eg, relighting, confirming the compatibility of our results with conventional rendering engines.
%
Our contributions are summarized as follows:
\begin{itemize}
    
    \item We design a simple yet effective dynamic weighting loss mechanism from the color decomposition for geometry reconstruction of challenging glossy objects.
    
    \item We propose a novel {\model} ({\mm}) to represent the global illumination and a material-aware cone-sampling method to effectively integrate \mm{} over BRDF lobes for high-fidelity material estimation.
    \item We constructed benchmarks (including both synthetic and real-world data) for the inverse rendering of challenging glossy objects with complex lighting interactions, and conducted extensive experiments to demonstrate the effectiveness of our method.
    
\end{itemize}

\begin{figure*}[t]
    \centering
    \makebox[\textwidth]{%
    \includegraphics[width=1.07\textwidth]{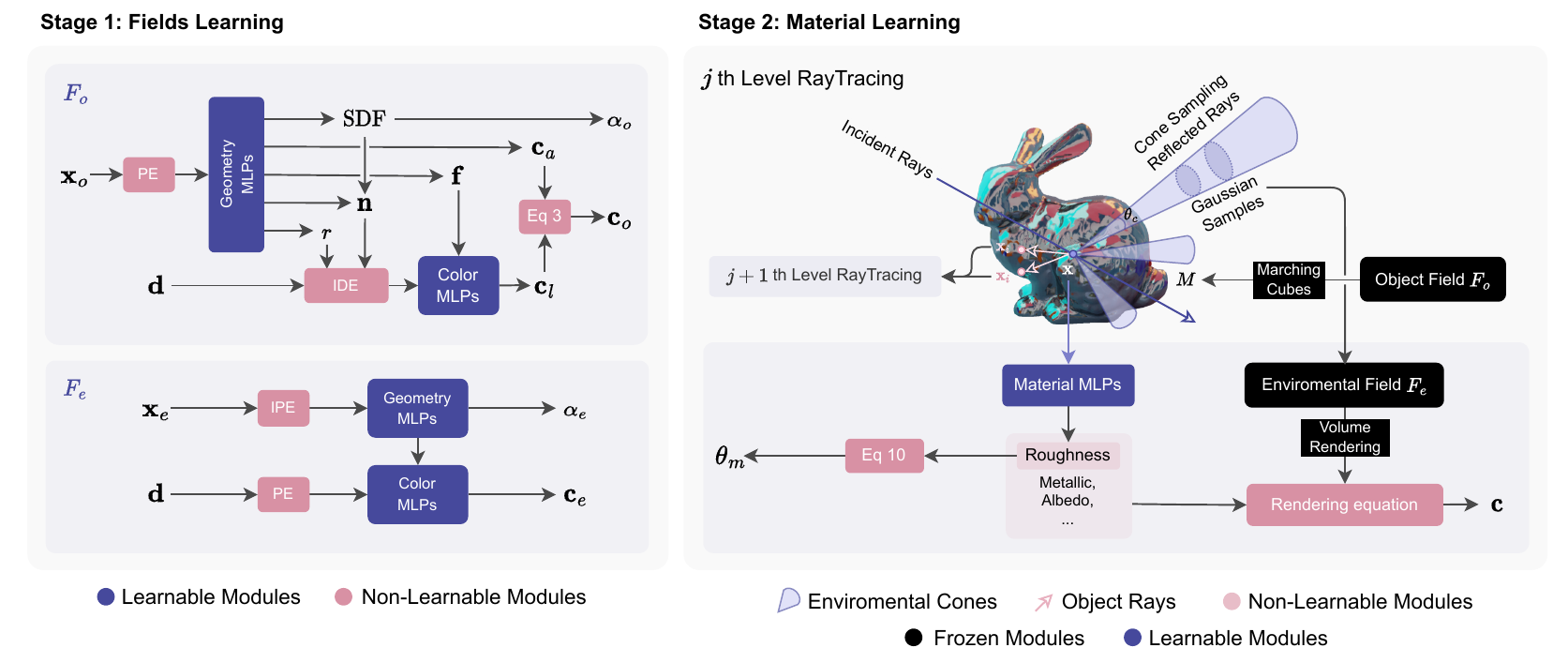}
}
\vspace{-2em}
    \caption{The detailed structure of our proposed method. Fields learning stage consists of an SDF-based object field and a Mip-NeRF as the environmental field. Based on them, we construct our {\model} via ray tracing and material-aware cone sampling method to represent the global illumination.
    }
\vspace{-1em}
    \label{fig:arch}
\end{figure*}
\section{Related Work}
\label{sec:related}

\noindent \textbf{Neural Radiance Fields.}
The introduction of Neural Radiance Fields (NeRF)~\cite{nerf} has marked a significant milestone in the field of 3D scene reconstruction, offering a new perspective on synthesizing novel views of complex scenes with details and realism. NeRF utilizes a fully connected neural network to model the volumetric scene radiance function. Building upon the foundation of NeRF, numerous works have sought to enhance its capabilities, addressing limitations and expanding its application range. Some methods~\cite{mipnerf, mipnerf360, trimipnerf} utilize enhanced cone sampling strategies rather than ray sampling, to address the aliasing problem and improve the captured details. Some methods~\cite{ngp, tensorf, plenoxel} focus on optimizing NeRF for faster training and inference times. Other methods~\cite{rawnerf, llnerf, deblur-nerf} improve NeRF for more robust training under degraded imaging conditions. These advancements collectively contribute to the evolution of NeRF, and solidify its position as a pivotal tool for 3D scene representation.

While NeRF models have demonstrated remarkable capabilities in synthesizing novel views, a key limitation lies in their inability to directly export reconstructed objects to rendering engines. A few NeRF-based inverse rendering approaches have been proposed to address this limitation.

\noindent \textbf{NeRF for Inverse Rendering.}
Leveraging the potential of NeRF for inverse rendering tasks has garnered substantial attention, aiming to retrieve the intrinsic properties, especially physically-based properties of objects or scenes from 2D images with camera poses. Specifically, inverse rendering of a target object in NeRF aims to reconstruct both the geometry structure and the material information of the object. In this realm, a number of methods have been proposed to unravel the complex interplay among geometry, material properties, and lighting.
High-quality inverse rendering results on real-world data are dependent on realistic rendering process simulation, which can be modeled by the framework of the rendering equation, to help decompose the physically based properties and reduce ambiguity. 

Previous methods mimic the rendering equation in different manners.
\cite{Zhang2021NeRFactor, refnerf} represent the early attempts in this direction, utilizing NeRF to infer the surface normal,  reflectance, and coarse lighting conditions simultaneously. Subsequently, some methods~\cite{nvdiffrec, nvdiffrecmc, neilf} try to improve geometry representation and propose a more accurate material estimating pipeline using Image-Based Lighting, facilitating a more robust and accurate inverse rendering process. They further improve the reconstruction quality by approximating the rendering equation with Monte Carlo sampling. Recently, some methods~\cite{nero} took a step further by independently estimating the geometry and material properties of the object, and incorporating occlusion-aware constraints into the NeuS-based inverse rendering framework, ensuring that the reconstructed geometry has fine details and material properties adhere to real-world physics. 

However, the existing methods either mainly utilize a 2D environmental map for lighting representation in the rendering equation, implicitly assuming that alllights over the scene come from infinity, or models direct lighting only~\cite{neai}. These assumptions often lead to less realistic renderings, especially in scenarios where light sources or other objects are in closer proximity to the target object. In contrast, our approach utilizes \model{} based on \fields{} for more realistic shape and material learning, overcoming this fundamental limitation.

\section{Method}

Our approach targets the inverse rendering of objects from calibrated multi-view images.
As our method is based on Neural Radiance Fields (NeRFs) and physically-based rendering (PBR), we start by briefly revisiting the relevant concepts in~\autoref{sec:method_pre}.
We then introduce our two-stage pipeline, which involves a fields learning stage (\autoref{sec:method_field}) for geometry reconstruction and environmental lighting learning, and a material learning stage (\autoref{sec:method_material}) for material estimation, as illustrated in~\autoref{fig:pipeline}.

\subsection{Preliminaries}
\label{sec:method_pre}


\noindent \textbf{Neural Radiance Fields (NeRF)} models the scene as a continuous function that maps a 5D vector (spatial location $\xx$ and viewing direction $\dd$) to a color $\cc = f_c (\xx, \dd)$ and a volume density $\sigma = f_d (\xx)$, where $f_c$ and $f_d$ are MLPs for predicting color and density, respectively.
%
%
While training, NeRF (which is parameterized as $\Theta_F$) casts camera rays for each pixel, and samples points or Gaussian samples~\cite{mipnerf} along the ray. The color of the ray is computed as:
\begin{equation}
L_\text{NeRF}(\Theta_F, \mathbf{r}) = \int_{t_n}^{t_f} T(t) f_d(\mathbf{r}(t)) f_c(\mathbf{r}(t), \mathbf{d}) \, dt \approx \sum_{i=1}^n w_i \cc_i, 
\label{eq:nerf}
\end{equation}
where \( \mathbf{r}(t) = \mathbf{o} + t\mathbf{d} \) is the parametric representation of the camera ray, \( T(t) \) is the accumulated transmittance along the ray, and $w_i$ is the weights for volume rendering. In practice, $T(t)$ can be defined as $T_i(t) = \prod_{j=1}^{i-1} (1 - \alpha_j)$.
%
%
Instead of the density, NeuS~\cite{neus} predicts Signed Distance Field (SDF) via $\sdf = f_\sdf(\xx)$, where the surface of the object is modeled by the zero-level set of SDF.
%
We represent the high-quality target object surfaces and environmental radiance based on both NeuS and NeRF.

\noindent \textbf{Rendering Equation}
aims to simulate the interaction of light and surfaces in a way that adheres to physical laws. 
Rendering equation, which is an integral equation describing the equilibrium of light in a scene, is the core of PBR. It is given by:
\begin{equation}
\reL (\xx, \dd) = \int_{\Omega} f_r(\xx, \dd_i, \dd) L_i(\xx, \dd_i) (n \cdot \dd_i) \, d\omega,
\label{eq:render}
\end{equation}
where \( \reL (\xx, \dd) \) is the outgoing radiance from point \( \xx \) in the view direction \( \dd \), \( L_i(\xx, \dd_i) \) is the incoming radiance from direction \( \dd_i \), \( f_r \) is the BRDF (Bidirectional Reflectance Distribution Function), \( n \) is the surface normal at point \( \xx \), and \( d\omega \) represents an infinitesimal solid angle.



Based on NeRF and PBR techniques, our approach improves the neural inverse rendering of glossy objects
by innovatively applying \fields{} and \model{} (\mm). 

\subsection{Fields Learning}
\label{sec:method_field}

Since simultaneously modeling geometry, lighting, and materials would lead to ambiguity, we construct
a two-stage pipeline to optimize the geometry and materials separately.
Our primary objective of the first stage is to simultaneously reconstruct the precise geometry of the target object and the environmental radiance field. 
This is the foundation for the subsequent material learning stage, which requires accurate light-surface intersection and surface normal.

Recent works~\cite{nero, nvdiffrecmc} have explored geometry reconstruction for glossy objects by incorporating physical priors like image-based rendering and split-sum approximations. However, they oversimplify the illumination as a 2D environmental map, which would cause suboptimal geometry, as shown in \autoref{sec:exp}. 
Although NeRO~\cite{nero} extends the environmental map to a directional function defined on the sphere, it may not capture the depth information of the scene, which struggles to reconstruct high-fidelity geometry for high-detailed objects in some cases.
%
To this end, we first propose to decouple the final color $\cc_o$ into albedo color $\cc_a$ and color modulated by the lighting $\cc_l$, as:
\begin{equation}
\cc_o = \cc_a \circ \cc_l,
\label{lr}
\end{equation}
\ryn{(in line with~\cite{refnerf, llnerf})}.
Based on the decomposed colors, we employ a dynamic weighting loss mechanism, inspired by~\cite{nerfw, refneus}, to strategically reduce the impact of highly uncertain reflective regions while amplifying the significance of diffuse areas \ryn{when computing the photometric loss}.
This mechanism ensures high-quality geometry reconstruction \ryn{of} glossy objects. Although it may distort the learned reflective color, it is inconsequential as we will estimate more accurate decomposed physical materials in the second stage. 
Unlike Ref-NeuS~\cite{refneus}, which
determines the loss weights for rays by correlating pixels across views,
we propose a simplier yet potent strategy, \ie, leveraging the discrepancy between albedo and final color to derive the weights:
\begin{equation}
    w_s = \min
    \left(\frac{1}{(\cc_o - \cc_a)^2 + \epsilon}, u\right),
\end{equation}
where $\epsilon$ denotes a small value. $u$ is a hyperparameter (set to $1.5$ by default) to cap the upper bound of the weights.
The intuition of our strategy lies \ryn{is}
that the discrepancy between albedo and final color is higher in reflective regions.

The field architectures in the first stage are illustrated in the left part of~\autoref{fig:arch}.
We represent the target object in a NeuS-based field $F_o$ and employ a neural radiance field $F_e$ to model the background environment.
During the volume rendering step, 
we divide samples along a ray into two sets with the border of the target object: object samples $\xx_o$ and environmental samples $\xx_e$.
%
Object samples $\xx_o$ are first featurized by the position encoding (PE)~\cite{nerf} and then fed into the Geometry MLPs to predict the SDF value, which is further mapped to opacity $\alpha_o$ as in NeuS~\cite{neus}, 
albedo $\cc_a$, roughness $r$, and a feature vector $\mathbf{f}$.
Next, we encode the view direction $\dd$, normal vector $\nn$ (derived from the SDF), and the roughness $r$ by  Integrated Directional Encoding (IDE)~\cite{refnerf}.
The IDE features and the feature vector $\mathbf{f}$ are then fed into the Color MLPs to predict the colors modulated by the lighting $\cc_l$.
The final colors $\cc_o$ for samples $\xx_o$ are produced by \autoref{lr} from the predicted $\cc_a$ and $\cc_l$.
On the other hand, environmental samples $\xx_e$ are mapped to opacity $\alpha_e$ and color $\cc_e$ by a Mip-NeRF~\cite{mipnerf} model.
Finally, the opacities ($\alpha_o$, $\alpha_e$) and colors ($\cc_o$, $\cc_e$) of object and environmental samples are rendered together into pixel color $\hat{\cc_p}$ by volume rendering as Eq.~\ref{eq:nerf}.
The whole fields are trained jointly by a weighted photometric loss:
\begin{equation}
    L_c = w_l \cdot |\cc_p - \hat{\cc}_p|,
\end{equation}
where $\hat{\cc}_p$ is the GT pixel color, and $w_l$ is the pixel-wise loss weight accumulated from $w_s$ along the rays via volume rendering. Besides the photometric loss, we also utilize a loss function to constrain the curvature of normal vectors. Refer to the Supplemental for more details.

\begin{figure*}[ht]
\setlength{\imw}{0.163\textwidth}
\renewcommand{\tabcolsep}{2pt}
\centering
\begin{tabular}{cccccc}
\includegraphics[width=\imw]{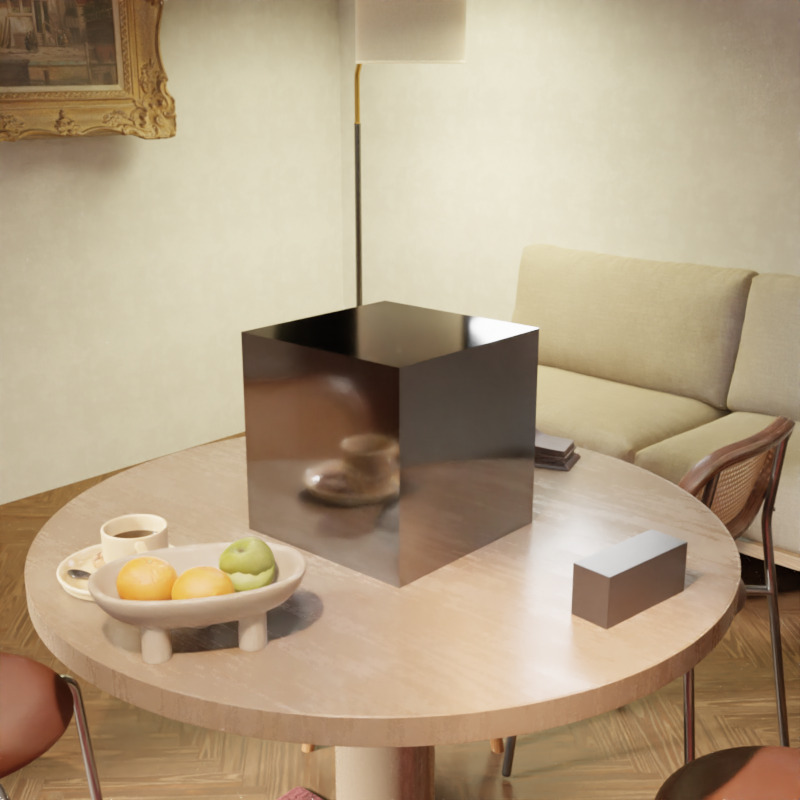} &
\includegraphics[width=\imw]{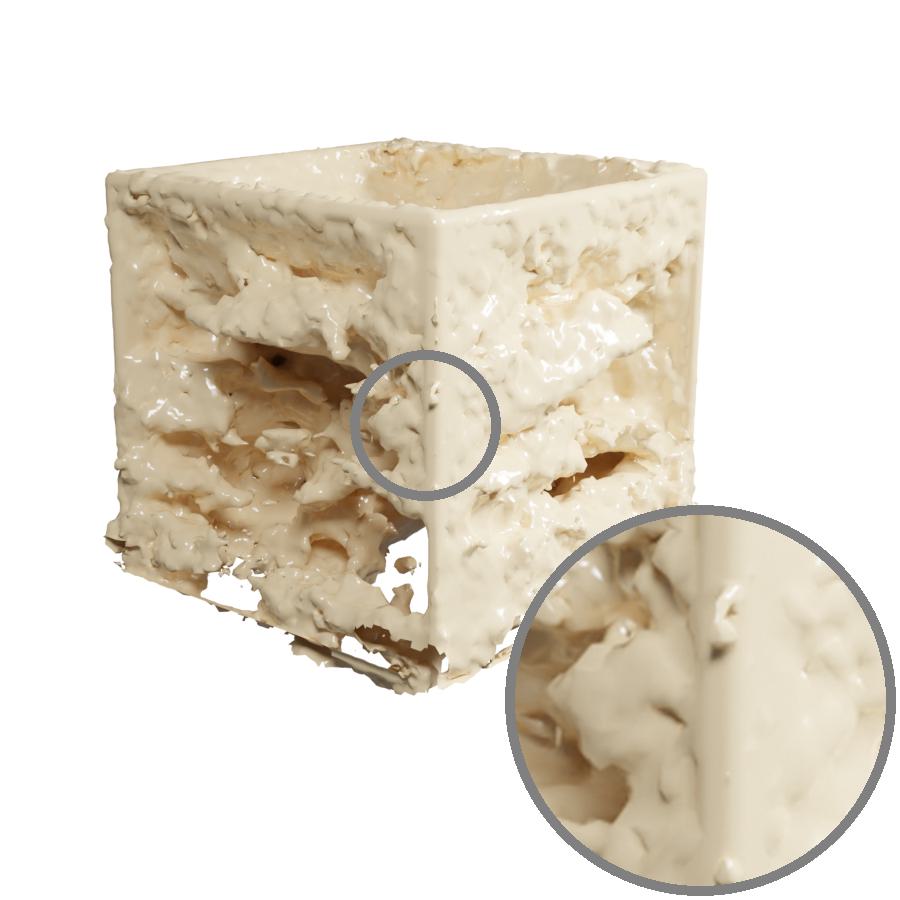} &
\includegraphics[width=\imw]{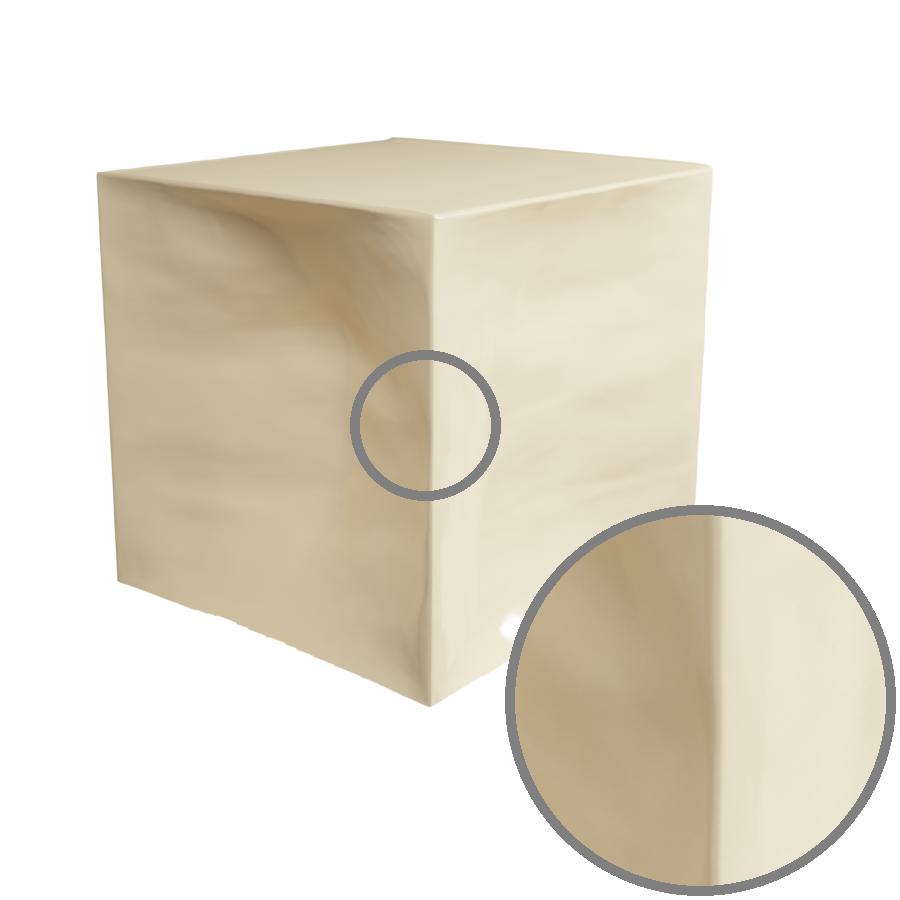} &
\includegraphics[width=\imw]{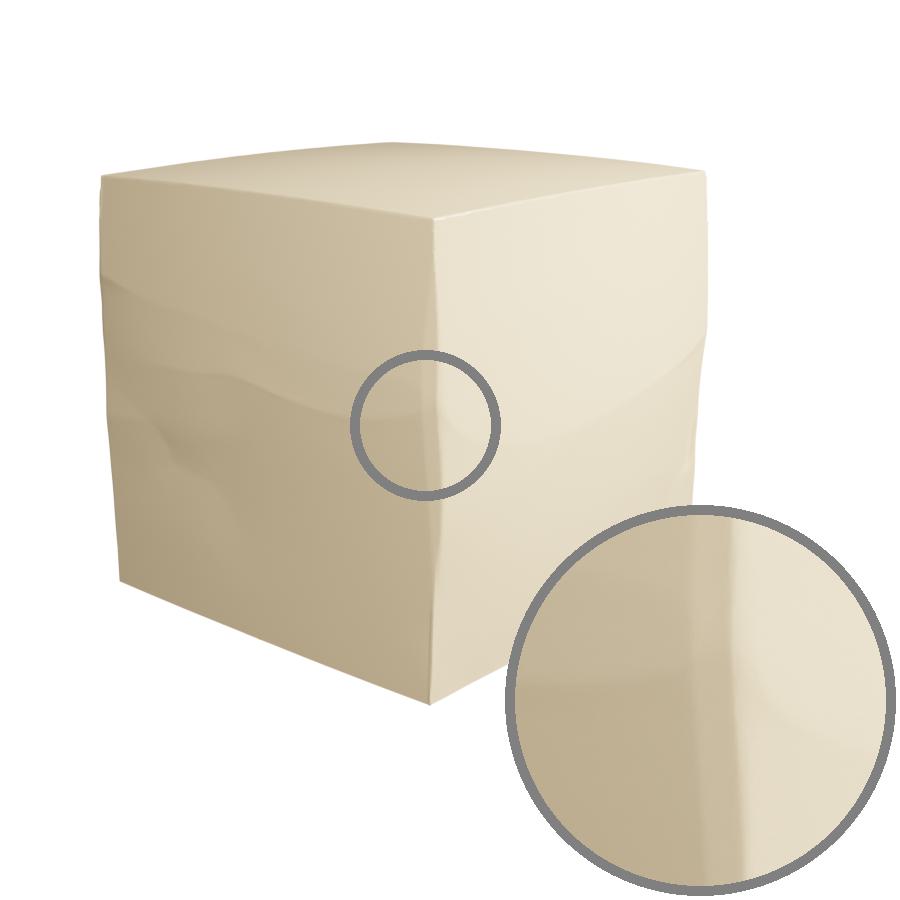} &
\includegraphics[width=\imw]{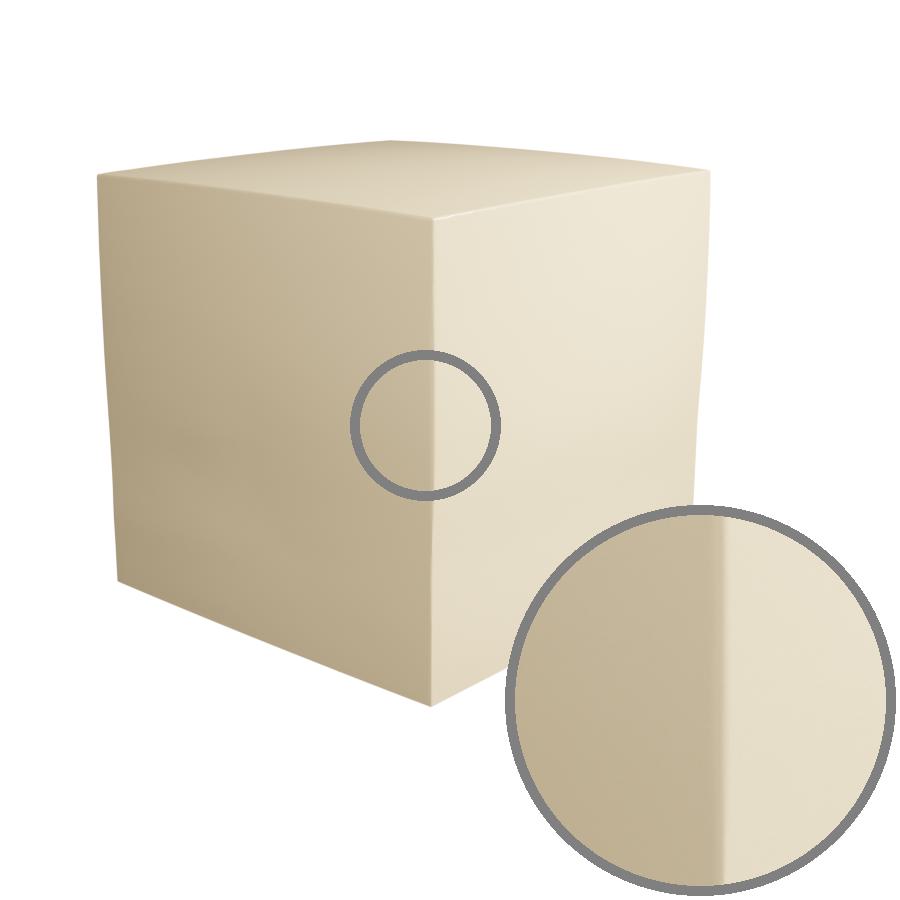} &
\includegraphics[width=\imw]{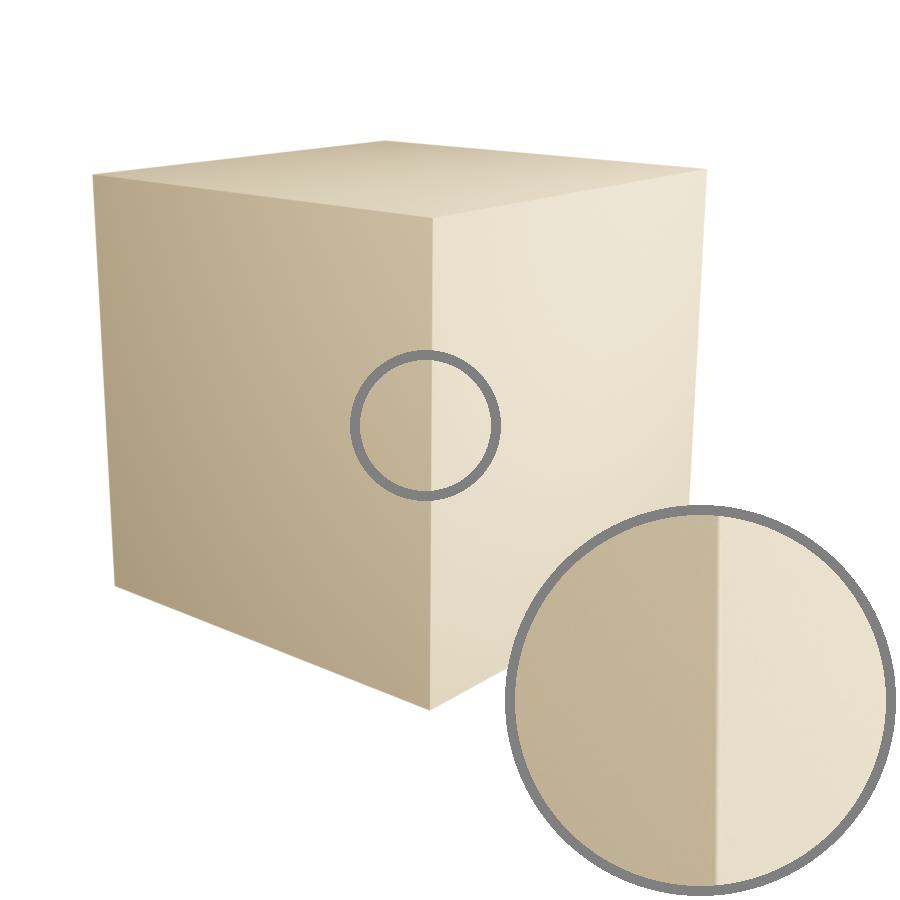} \\


\includegraphics[width=\imw]{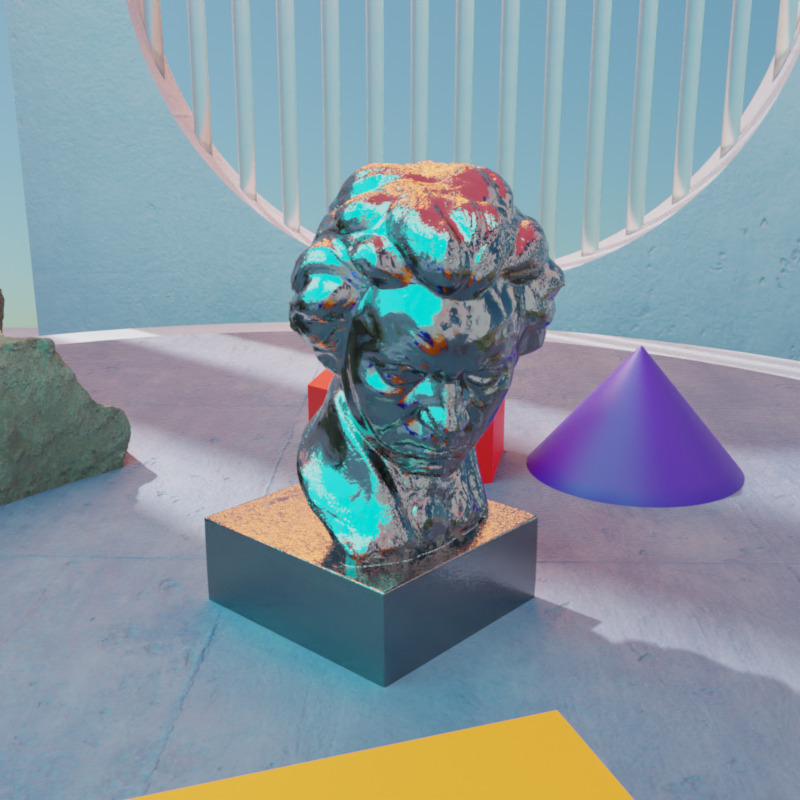} &
\includegraphics[width=\imw]{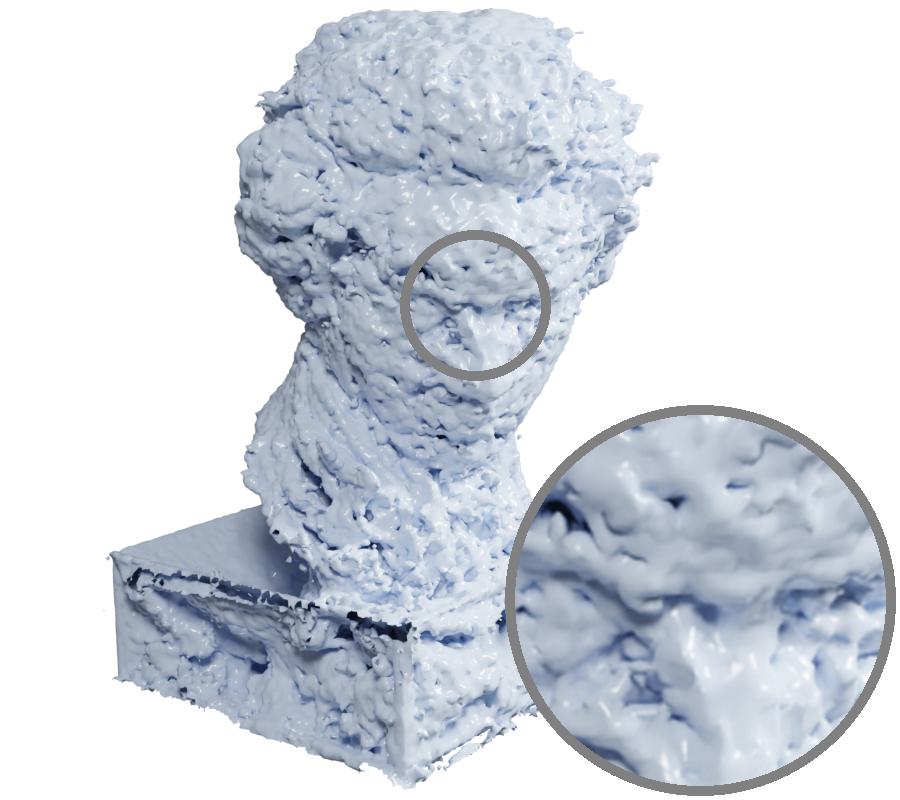} &
\includegraphics[width=\imw]{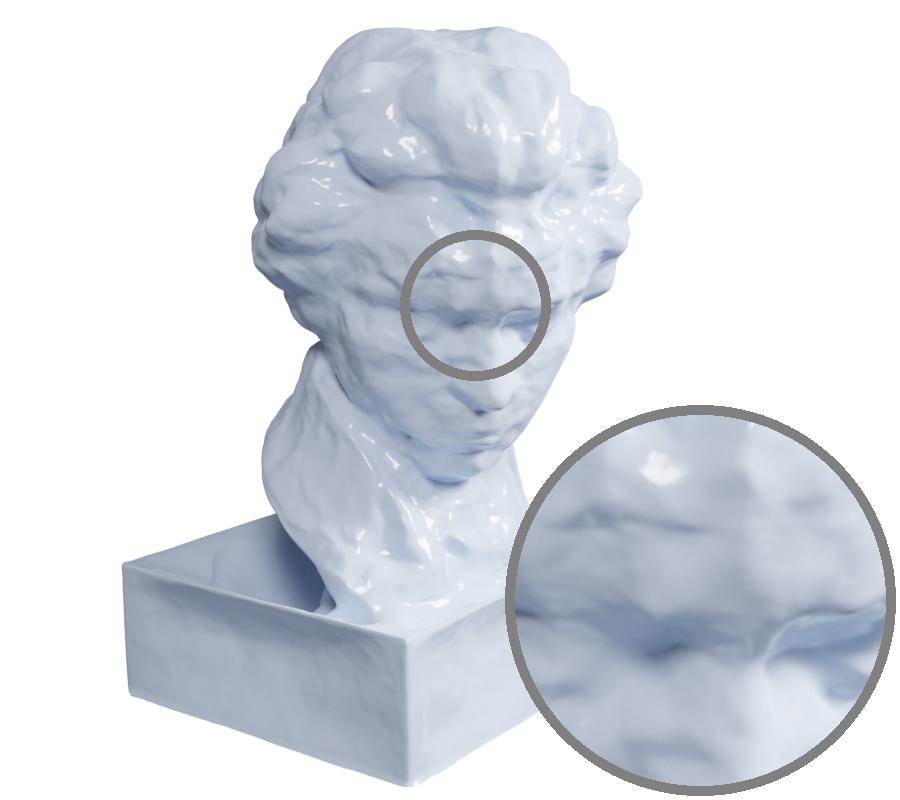} &
\includegraphics[width=\imw]{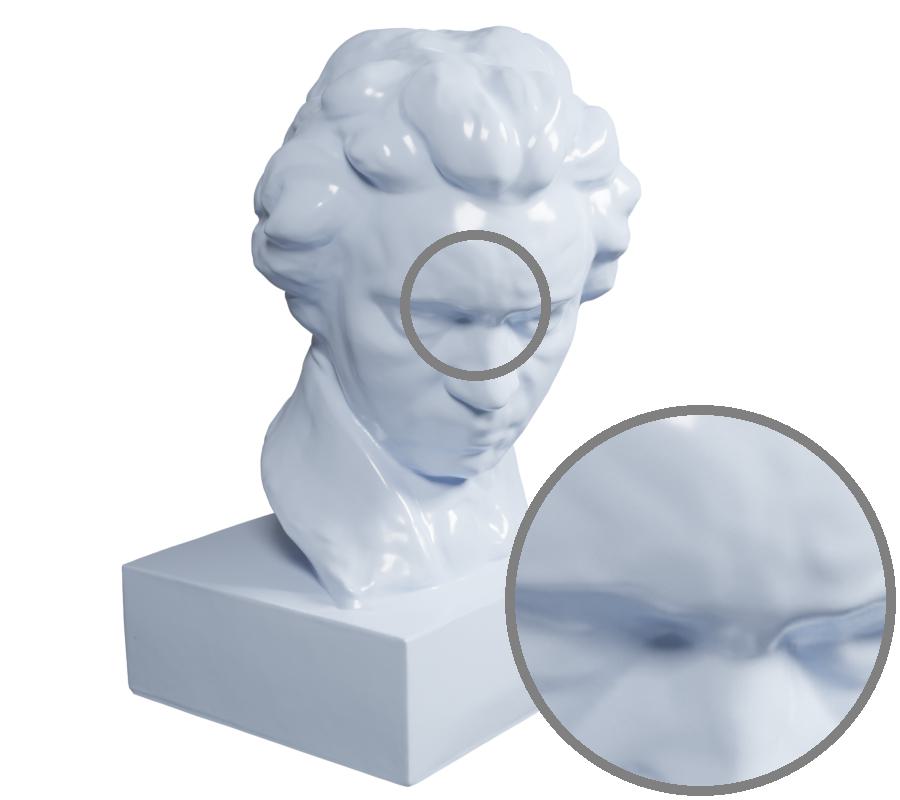} &
\includegraphics[width=\imw]{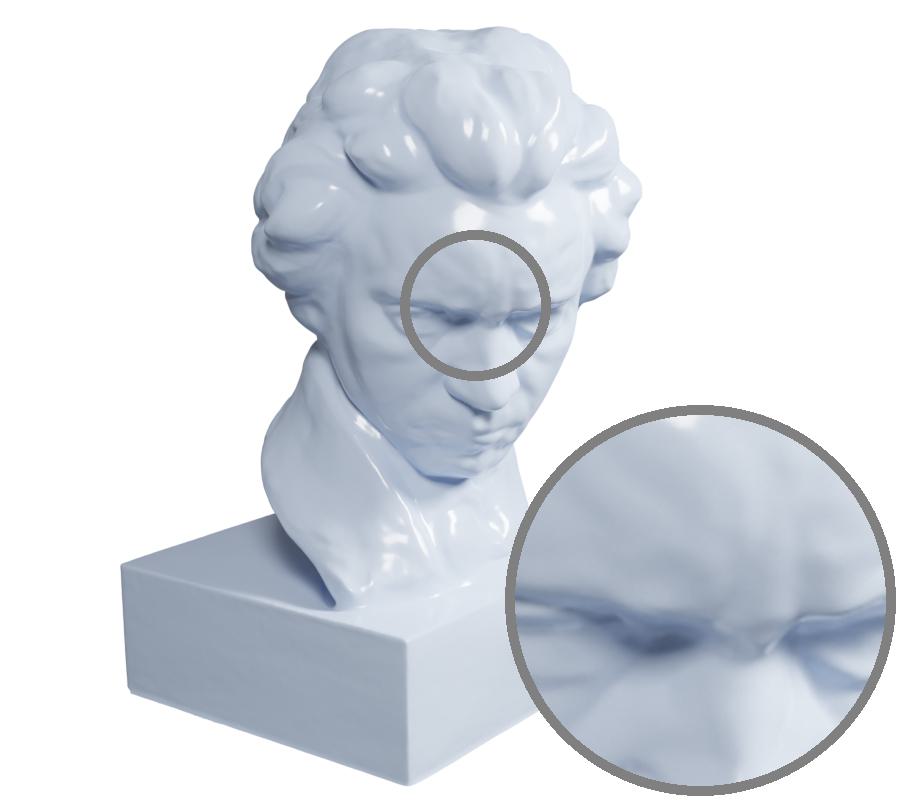} &
\includegraphics[width=\imw]{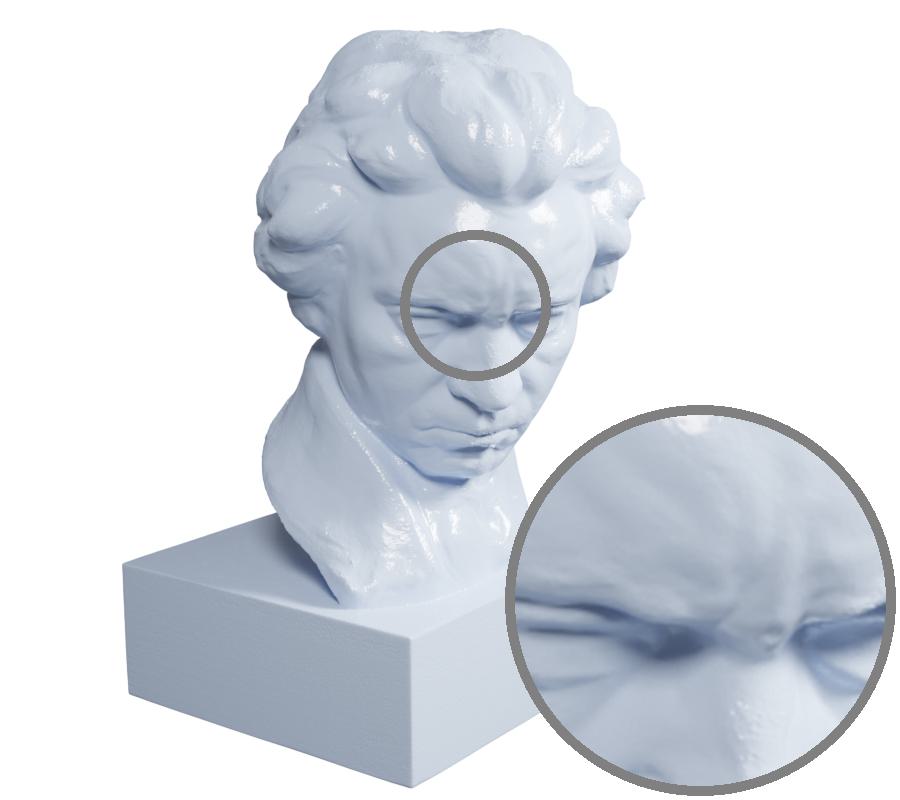} \\

\includegraphics[width=\imw]{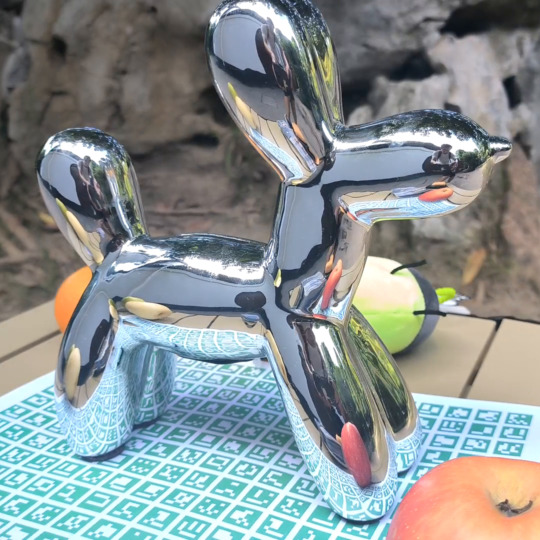} &
\includegraphics[width=\imw]{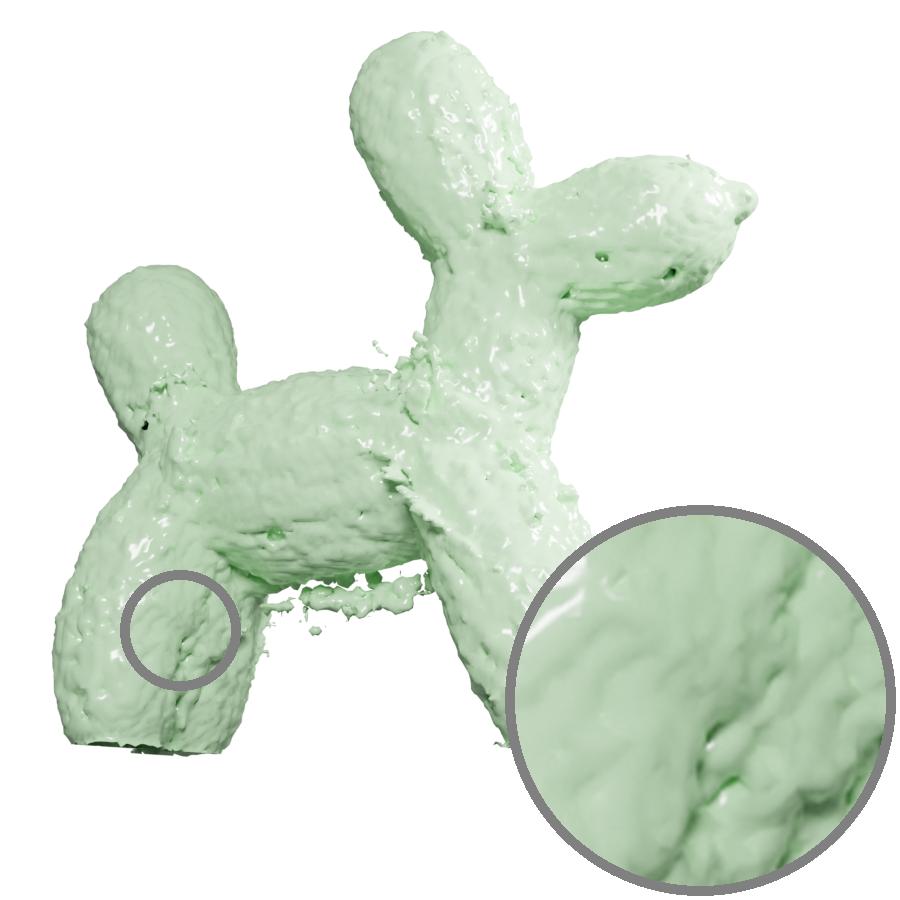} &
\includegraphics[width=\imw]{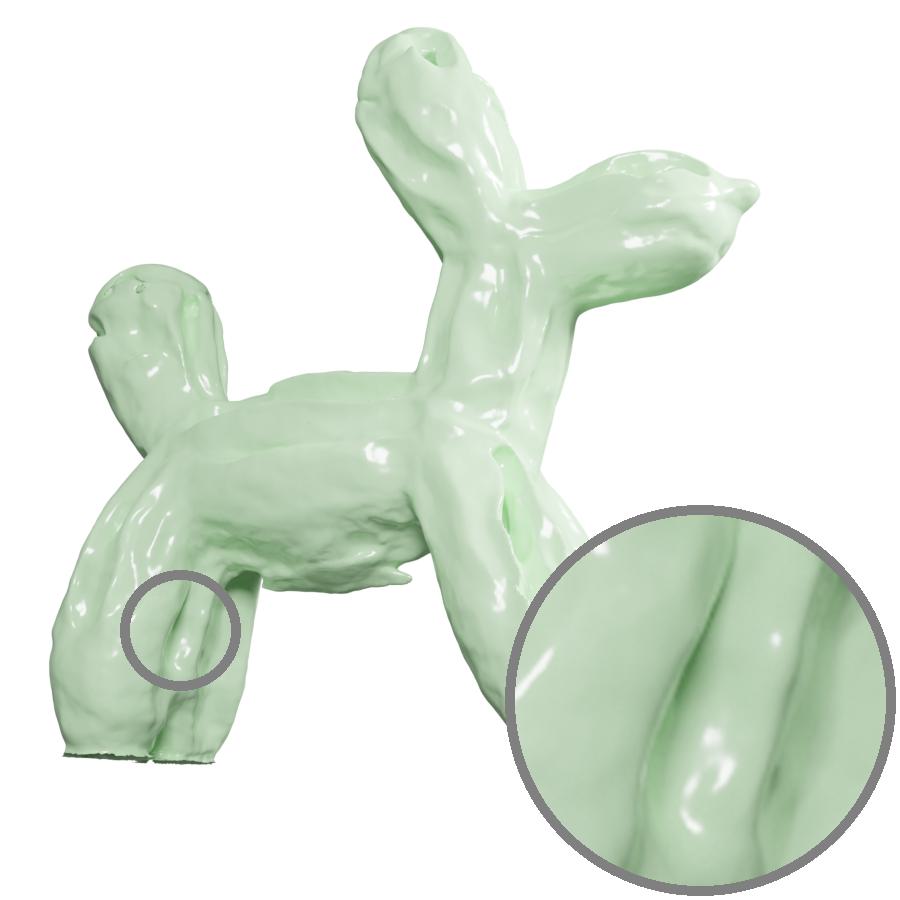} &
\includegraphics[width=\imw]{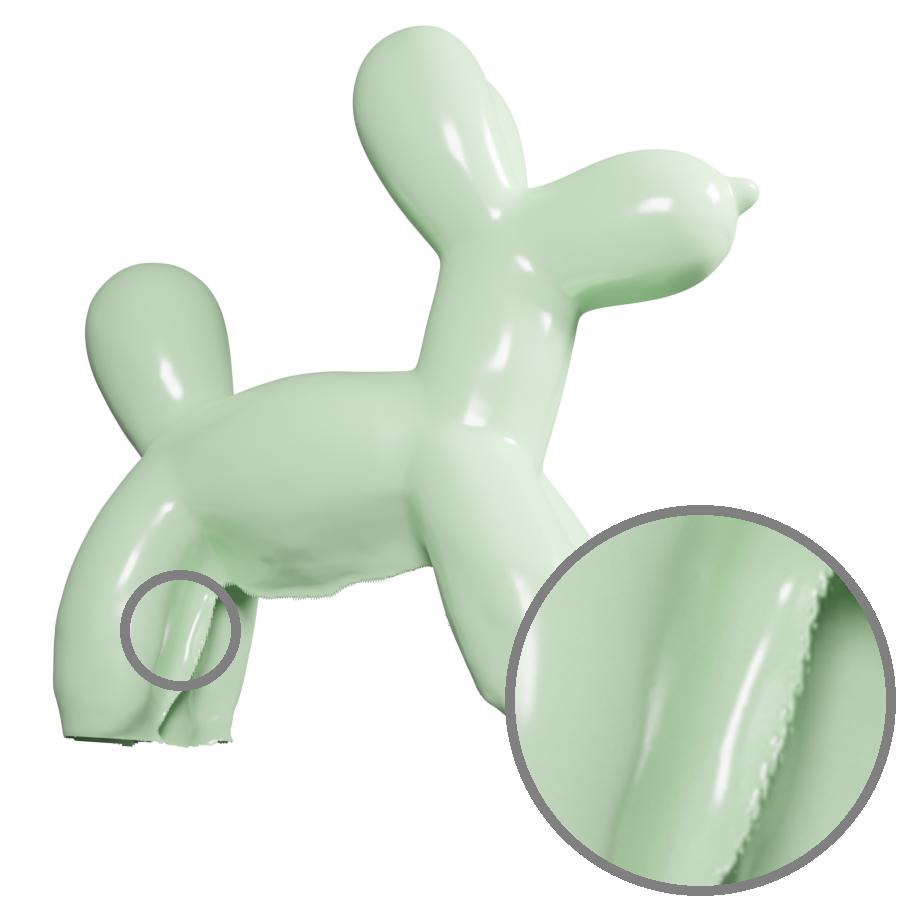} &
\includegraphics[width=\imw]{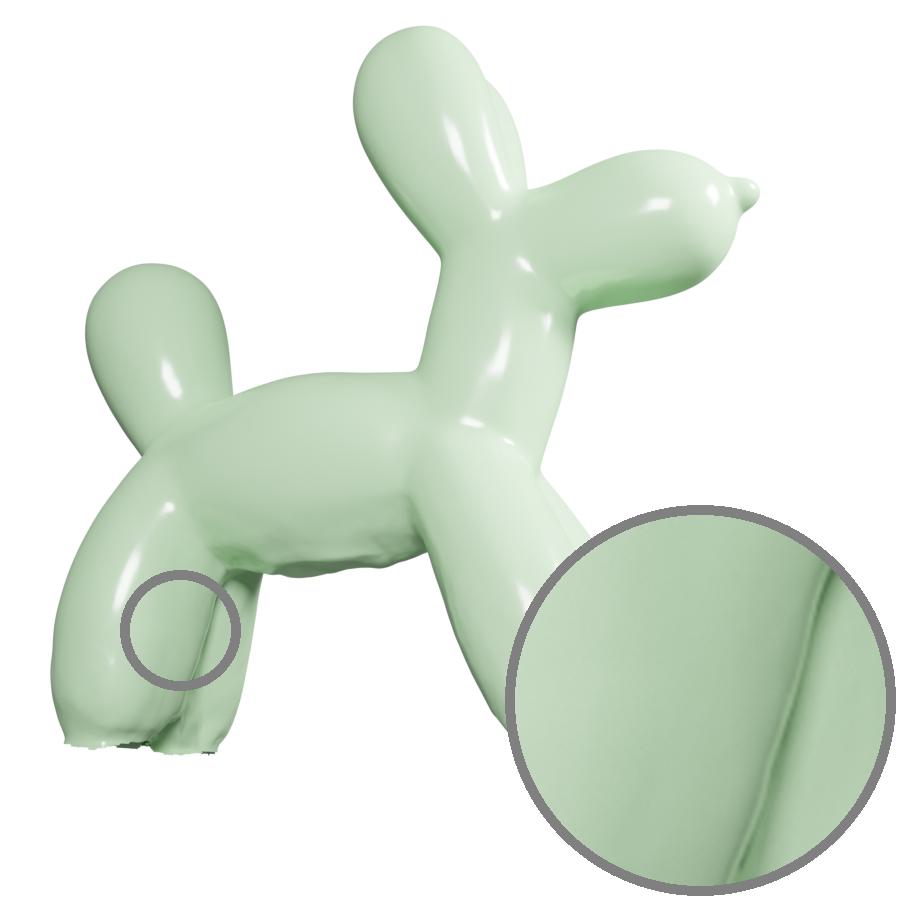} &
\includegraphics[width=\imw]{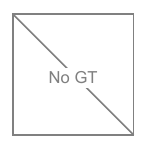} \\

{\footnotesize Input Sample} &
{\footnotesize Nerfacto~\cite{nerfstudio}} &
{\footnotesize NeuS~\cite{neus}}&
{\footnotesize NeRO~\cite{nero}}&
{\footnotesize Ours} &
{\footnotesize GT} \\

\end{tabular}
\vspace{-4mm}
\caption{Comparison of the geometry reconstruction among cutting-edge methods and ours. 
For each method, we utilize marching cubes to extract the triangle meshes for comparison.}
\vspace{-1em}
\label{fig:shape-cmp}
\end{figure*}

\begin{figure}[ht]
    \centering
    \includegraphics[width=1.0\linewidth]{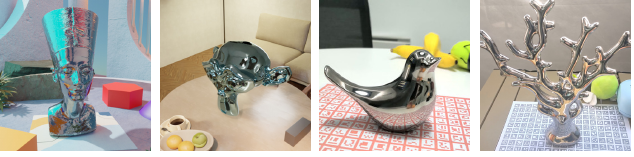}
    \vspace{-6mm}
    \caption{Samples from the proposed dataset. The first two examples are synthetic and the last two samples are real-world captured.}
    \label{fig:ds}
    \vspace{-3mm}
\end{figure}

\subsection{Material Learning}
\label{sec:method_material}

After obtaining the precise geometry of the target object, our goal for the second stage is to estimate the physically-based materials of the object.
To achieve this, we adopt a more comprehensive rendering process, \ie, ray tracing, to evaluate the rendering equation in \autoref{eq:render}.
%
%
We can employ the marching cubes algorithm~\cite{marching-cubes} to extract the triangle mesh of the object by efficiently determining the ray-surface intersection $\xx$ and surface normal $\nn$.
Thus, the key to estimating the materials is to faithfully represent the global illumination $L_i$ in the rendering equation.

\paragraph{Neural Plenoptic Function.}
Instead of simplifying the representation of the lighting as a 2D environment map, we propose a 5D neural plenoptic function (\mm), symbolized by $f_p(\xx, \dd)$, to represent the color of the light observed at the spatial location $\xx$ with the direction $\dd$, \ie $L_i = f_p(\xx, \dd_i)$.
However, it is challenging to directly learn \mm{} due to the high dimensionality.
Because of the superiority of NeRF in representing the radiance field, we propose to construct \mm{} from the pre-trained environmental radiance fields $F_e$ in the first stage.
Specifically, we start by modeling the intersection between the incoming light $\mathbf{r}_i(t) = \xx + t\dd_i$ and the object mesh $M$ as a function $I(\mathbf{r}_i, M)$ to determine the point of contact $\xx_i$.
If there are no intersections, which means that the incoming light is a direct light from the environment, we utilize \autoref{eq:nerf} with the pretrained $F_e$ to define the lighting color.
We also utilize a simple MLP to learn the residual value of $L_\text{NeRF}$, permitting the model to learn the distant lighting that is not captured by training images.
%
Conversely, if the intersection point $\xx_i$ exists, we employ a ray-tracing approach based on \autoref{eq:render} to obtain the color of the incoming light as it is an indirect light in this situation.
%
The plenoptic function is defined as:
%
\begin{subequations}
\begin{empheq}[left={f_p(\xx,\dd_i)=}\empheqlbrace]{align}
    & \reL (\xx_i, \dd_i), & \text{ if } \exists \; \xx_i, \label{eq:plen-obj} \\
    & L_\text{NeRF}\left(\Theta_{F_e}, \mathbf{r}_i(t)\right), & \text{ otherwise}, \label{eq:plen-nerf}
\end{empheq}
\end{subequations}
where $L$ is a discretized rendering equation discussed next.



%


\paragraph{Ray Tracing.}
\label{sec:tracing}
The discretized rendering equation $L$ is derived via the ray tracing algorithm as:
\begin{equation}
\reL (\xx, \dd) = \sum_{k=1}^m f_r(\xx, \dd_k, \dd) f_p(\xx, \dd_k) (\nn \cdot \dd_k),
\label{eq:tracing}
\end{equation}
%
%
where $m$ is the number of sampled incoming rays per intersection point.
Note that the formulation of $f_p$ in \autoref{eq:plen-obj} also incorporates the computation of $L$. Thus, a recursive ray tracing process is constructed. 
The ray tracing algorithm unfolds in $N$ levels. 
At each level, it samples a set of incoming lights emitted from the current shading point $\xx$, where the color of each light is given using \autoref{eq:plen-nerf} if no intersections are found with the object, and the tracing process is terminated. Otherwise, the ray tracing delves deeper into the next level, permitting the light to do an additional bounce.
Upon the $N$-th level tracing, if there still exists an intersection point with the object mesh, we employ an MLP to predict the lighting color: $L^{(N)}(\xx, \dd) = \text{MLP}(\xx, \dd)$, thus concluding the journey of the light.

\paragraph{Material-Aware Cone Sampling.}
\label{sec:cone}
Representing light via the proposed 5D neural plenoptic function offers a fidelity improvement. However, directly applying it is not practical, as it introduces a significant computational cost in the ray marching of NeRF, which demands sampling along rays to get the color information for each individual light. The complexity is further increased when doing important ray sampling for ray tracing within the BRDF lobe to satisfy the rendering equation. 

To address this challenge, we introduce a material-aware cone sampling technique. In the first stage, we adopt a Mip-NeRF as our environmental field, which samples cones instead of rays like a vanilla NeRF. The introduction of Mip-NeRF is informed by the innate congruence of its cone sampling with the BRDF lobe-centric importance ray sampling. During the training of Mip-NeRF in the fields learning stage, the pre-integrated Gaussian samples are employed through Integral Positional Encoding (IPE), yielding better rendering results than a vanilla NeRF without incurring obvious extra costs. During the fields learning stage, the environmental field $F_e$ is trained as a cone pre-filterer supervised by the training images. 

In the material learning stage, we fix $F_e$ and sample cones from the BRDF lobe. Because of the correlation between surface roughness and GGX distribution, we derive the cone angle directly from the predicted roughness. Subsequent ray marching of $F_e$ yields the pre-filtered color for the incoming light of each integral component in the rendering equation.
Specifically, considering the roughness parameter $r$ at a shading point, we adopt the GGX distribution function $D(\mathbf{m})$ to describe the probability distribution of the microfacet normals $\mathbf{m}$. From this, we have the probability density function (PDF) for the azimuth angle $\phi$ and elevation angle $\theta$~\cite{microfacet-sampler}:
\begin{equation}
    p_m(\theta, \phi) = \frac{r^4 \cos\theta \sin\theta}{\pi ((r^4 - 1) \cos^2\theta + 1)^2}.
\end{equation}

Our aim is to determine $\theta_m$, which bounds the range of sampling the orientation of the microfacet normal $\mathbf{m}$, allowing the BRDF lobe to capture a predefined portion $\beta$ of the light energy over the hemisphere. For practical purposes, we select $\beta=0.9$ to encompass 90\% of the radiance energy. This leads us to establish the cumulative distribution function (CDF) $P_m$ and to resolve the following equation:
\begin{equation}
     P_m(\theta_m) = \frac{r^4}{\cos^2\theta_m(r^4-1)^2 + (r^4-1)} - \frac{1}{r^4-1}, 
\end{equation}
wherein the solution of $P_m = \beta$ is presented as:
\begin{equation}
\theta_m = \arctan \left(r^2 \sqrt{\frac{\beta}{1 - \beta}} \right). 
\end{equation}

The angle $\theta_m$, indicative of the spatial extent of the BRDF lobe, is thus directly correlated with the surface roughness. Consequently, we can sample the cone with the apex angle $\theta_c$ using $\theta_m$ via a simple math transform, a method without the need for learnable parameters. The details of the derivation can be found in the Supplemental.

Leveraging the pre-filtered lighting significantly reduces the number of lighting samples. Typically, employing a GGX distribution-based important sampling requires about 256 diffuse and 128 specular rays to get satisfactory results. In contrast, the proposed method achieves competitive results with merely 8 diffuse and 4 specular cones.

\begin{figure*}[ht]
\setlength{\imw}{0.16\textwidth}
\renewcommand{\tabcolsep}{2pt}
\centering
  \includegraphics[width=1.0\linewidth]{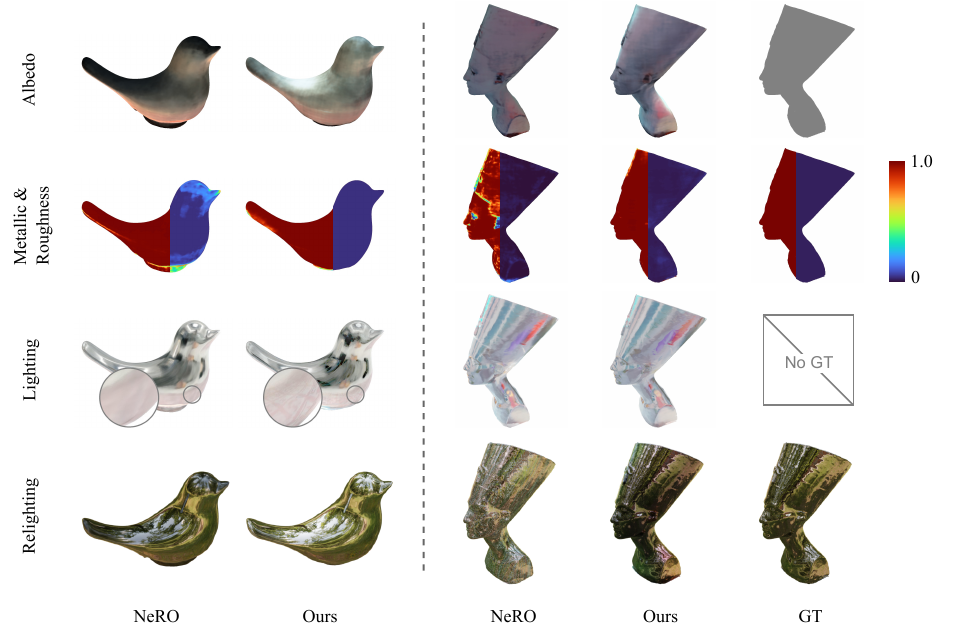}%

%

\vspace{-3mm}
\caption{Comparison on the material estimation results. We display one real-world data (bird) and one synthetic data (Nefertiti) with GT for visual comparison. In the images of metallic \& roughness row, metallic is displayed on the left half and roughness on the right half.
}
\label{fig:mat-cmp}
\end{figure*}

\section{Experiments and Analysis}
\label{sec:exp}

In this section, we detail the experiments conducted to validate the efficacy of our proposed method. We assess our method's performance in both geometry and material reconstruction, and demonstrate its practical applications.

\begin{figure*}[h]
    \centering
    \includegraphics[width=1.0\linewidth]{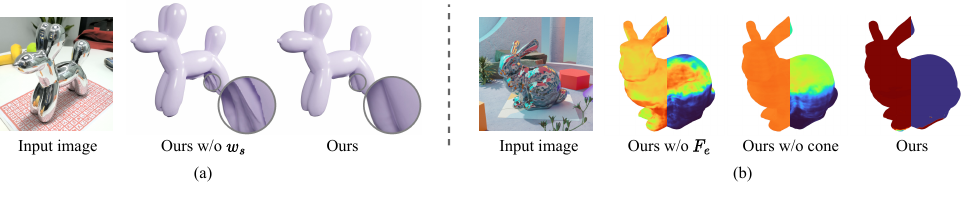}
    \vspace{-8mm}
    \caption{Ablation study results showcasing the impact of key components on inverse rendering quality for both stages. 
    The absence of dynamic weighting (Ours w/o $w_s$) leads to less smooth surface reconstruction, while the lack of environmental field (Ours w/o $F_e$) or material-aware cone sampling method (Ours w/o cone) diminishes material fidelity. In each image of Figure (b), metallic is displayed on the left half and roughness on the right half.
    }
    \label{fig:ablation}
\end{figure*}

\noindent{\bf Dataset.}
Our dataset is constructed to benchmark the inverse rendering of challenging glossy objects with diverse lighting interactions with nearby objects. 
It comprises 20 scenes in total: 10 real-world scenes captured through photographing glossy objects under varying lighting scenarios to produce multi-view images and 10 synthetic scenes. The synthetic scenes are crafted with glossy objects and manually designed background environments, and the multi-view images are rendered using  
photo-realistic path-tracing
rendering engine. A distinctive aspect of our synthetic dataset is that we did not use environmental maps as the background directly, which is commonly used in prior works. Instead, we focus on more realistic settings where the objects are situated within tangible environments consisting of other objects. Samples from the proposed dataset are shown in \autoref{fig:ds}. 

\noindent{\bf Geometry Evaluations.}
To evaluate the quality of our method in geometry reconstruction, we compare our
mesh outputs against those generated by several 
cutting-edge approaches,
including Nerfacto~\cite{nerfstudio}, NeuS~\cite{neus}, and NeRO~\cite{nero}.

\begin{table*}[ht]
\centering


\begin{tabular}{l|cccc|cc}
	\toprule
	                                              &
	\multicolumn{4}{c|}{Geometry Comparison (CD$\downarrow$)} &
	\multicolumn{2}{c}{Roughness / Metallic / Albedo Comparison (MSE$\downarrow$)} \\
	\cmidrule(lr){2-7}
	                                              &
	Nerfacto~\cite{nerfstudio}                                      &
	NeuS~\cite{neus}                                          &
	NeRO~\cite{nero}                                          &
	Ours                                          &
	NeRO~\cite{nero}                                          &
	Ours                                                               \\ \midrule
	Bunny                                         &
	0.05498                                       &
	0.00852                                       &
	\wcell{0.00153}                &
	\gcell{0.00147}                 &
	\gcell{0.002} / 0.022 / 0.044                         &
	\gcell{0.002} / \gcell{0.016} / \gcell{0.022}                                              \\
	Box                                           &
	0.07028                                       &
	0.03339                                       &
	\wcell{0.00412}                &
	\gcell{0.00135}                 &
	0.003 / 0.068 / \gcell{0.056}                         &
	\gcell{0.001} / \gcell{0.019} / 0.067                                              \\
	Beethoven                                     &
	0.04038                                       &
	0.01706                                       &
	\wcell{0.00197}                &
	\gcell{0.00146}                 &
	0.007 / 0.030 / 0.031                               &
	\gcell{0.004} / \gcell{0.026} / \gcell{0.027}                                                    \\
	Suzanne                                       &
	0.05753                                       &
	0.00754                                       &
	\wcell{0.00264}                &
	\gcell{0.00227}                 &
	0.004 / 0.030 / \gcell{0.024}                         &
	\gcell{0.001} / \gcell{0.022} / 0.029                                              \\
	Nefertiti                                     &
	0.05299                                       &
	0.01053                                       &
	\wcell{0.00587}                &
	\gcell{0.00167}                 &
	0.020 / 0.103 / 0.040                         &
	\gcell{0.008} / \gcell{0.021} / \gcell{0.035}                                              \\ \midrule
	Avg.                                          &
	0.05523                                       &
	0.01541                                       &
	\wcell{0.00322}                &
	\gcell{0.00164}                 &
	0.007 / 0.051 / 0.039                               &
	\gcell{0.003} / \gcell{0.020} / \gcell{0.036}                                                    \\ \bottomrule
\end{tabular}

\caption{Quantitative comparison results. We compare the Chamfer Distance (CD) between the meshes from our method and those from other methods for the synthetic objects with GT on the left, and compare the MSE between the materials from of our methods and those from other methods on the right. 
}
\label{tab:cmp}
\end{table*}



The visual comparison results are displayed in \autoref{fig:shape-cmp}, which provides a side-by-side comparison of the reconstructed objects. 
We also compare the Chamfer Distance (CD) with other methods on the synthetic scenes with GT meshes, as shown in \autoref{tab:cmp}. 
Our method consistently outperforms the others across all tested scenes, achieving the lowest average CD and better visual quality with better geometric reconstruction fidelity.
The comparison of the geometry reconstruction highlights
the strengths of our proposed 
dynamic weighting loss mechanism from the decoupled colors in the first stage.
The performance of our method in reconstructing accurate and detailed geometries lays the groundwork for the subsequent material property estimation and their applications of seamlessly integrated into the rendering engine.

\noindent{\bf Material Evaluations.}
Moving beyond geometry, we compare the material estimation results produced by our method with 
NeRO~\cite{nero}.
Given our synthetic data subset, we possess the ground truth values for albedo, roughness, and metallic maps, allowing for a direct quantitative assessment via Mean Squared Error (MSE) as illustrated in \autoref{tab:cmp}. For visual comparison, we present two samples to validate compared methods and GT in \autoref{fig:mat-cmp}. 
Both the quantitative and qualitative results reveal the competitive performances of our method in reconstructing materials with high fidelity, where our proposed \model{} is able to predict a more precise and smooth roughness, metallic, and slightly better albedo compared with the existing methods.



\noindent{\bf Ablation Study.}
To assess the contribution of different components in our model, we perform an ablation study on the two stages of our method. 
In the first stage, we study the effect of the proposed dynamic weighting method. We compare the result of our full model with that of the variant where the proposed dynamic weighting scheme is excluded (\ie, Ours w/o $w_s$). In the second stage, we study the results of our model in material estimation (1) by replacing the environmental field with an MLP to predict environmental lighting (\ie, Ours w/o $F_e$), and (2) by replacing the material-aware cone sampling mechanism by a fixed number of rays (\ie, Ours w/o cone).
The ablation results on two samples are shown in \autoref{fig:ablation}.

We can see that by omitting the dynamic weighting strategy, which is crucial in balancing the influence of uncertain regions, we observe a decrease in the model's ability to reconstruct a smooth surface for the target object.
Without the environmental field or material-aware cone sampling method for lighting representation, it leads to a degradation in material fidelity and introduces more ambiguity.
Overall, we have demonstrated through the ablation study the importance of different parts of our method on the quality of the final inverse rendering results.

\paragraph{Limitation.} 
While our method has been shown to make advancements in physically-based inverse object rendering, it is important to acknowledge certain limitations inherent in our current method.

One of the primary constraints of our approach is that the material learning stage is heavily reliant on the quality of the reconstructed geometry in the field learning stage. This dependency implies that the inaccuracies in the recovered geometry may adversely affect the material estimation process. If the geometry reconstructed in the first stage is not precise, 
it could lead to suboptimal material estimations in the subsequent stage.

Besides, although our method reduces the ambiguities typically present in inverse rendering tasks. we observe that the decomposition between lighting, geometry, and material properties is not entirely unambiguous. For example, there are scenarios where the color affected by the geometric structure is inaccurately incorporated into the albedo. This suggests an inherent complexity in perfectly disentangling the contributing factors to the final rendered image even with an explicit representation of lighting. 

In future works, addressing these limitations will be crucial to further enhance the robustness and accuracy of our neural inverse rendering method. Potential improvements may include introducing more sophisticated constraints for geometry-material interdependence and refining the decomposition process to minimize ambiguities.

\section{Conclusion}

In this paper, we have presented a novel physically-based inverse rendering method for glossy objects, with the introduction of \MODEL{} (\mm) based on NeRFs. Our method addresses the limitations of the dependency on the simplified lighting representation in previous NeRF-based inverse rendering approaches.
It is formulated as a two-stage model. The fields learning stage enhances the accuracy of 3D geometry reconstruction, especially for glossy objects under complex lighting. In the material learning stage, \mm{} employs a 5D neural plenoptic function for lighting representation based on the object field and environmental field, leading to higher-fidelity material estimation and inverse rendering. Our proposed material-aware cone sampling strategy further improves the efficiency of material learning. Experiments on real-world and synthetic datasets demonstrate the superior performance of our method.

\section*{Acknowledgements}
This work is partly supported by a GRF grant from the Research Grants Council of Hong Kong (Ref.: 11205620).

{
    \small
    \bibliographystyle{ieeenat_fullname}
    \bibliography{main}
}
\end{document}